\documentclass[11pt]{article}

\usepackage{acl}

\usepackage{times}
\usepackage{latexsym}
\usepackage[T1]{fontenc}
\usepackage[utf8]{inputenc}
\usepackage{microtype}
\usepackage{inconsolata}
\usepackage{graphicx}
\usepackage{amsmath}
\usepackage{amssymb}
\usepackage{booktabs}
\usepackage{colortbl}
\usepackage{enumitem}
\usepackage{multirow}
\usepackage{algorithm}
\usepackage{algpseudocode}
\usepackage{tabularx}

\title{ACE-SQL: Adaptive Co-Optimization via Empirical \\ Credit Assignment for Text-to-SQL}

\author{
Xiaobing Chen\textsuperscript{1,2}\thanks{Equal contribution.},
Ai Jian\textsuperscript{3}\footnotemark[1],
Eryu Guo\textsuperscript{3},
Zhiqi Pang\textsuperscript{1}\thanks{Corresponding author.} \\
\textsuperscript{1}Harbin Engineering University, Harbin, China \\
\textsuperscript{2}Harbin Institute of Technology, Harbin, China \\
\textsuperscript{3}Beijing University of Posts and Telecommunications, Beijing, China \\
\texttt{xbchen@stu.hit.edu.cn},
\texttt{zqpang98@hrbeu.edu.cn}
}

\begin{document}
\maketitle

\begin{abstract}
Text-to-SQL maps natural language questions to executable SQL queries.
Modern databases often contain large and complex schemas, making schema linking a critical step for accurate SQL generation.
Existing methods either rely on full-schema generation, which leaves schema linking implicit within a large search space, or use a separate retriever trained with static gold-column supervision, whose targets may be suboptimal for the current generator policy.
To address this issue, we propose \textbf{A}daptive \textbf{C}o-optimization via \textbf{E}mpirical Credit Assignment for Text-to-\textbf{SQL} (\textbf{ACE-SQL}), a reinforcement learning (RL) framework that jointly optimizes schema retrieval and SQL generation under execution feedback.
ACE-SQL constructs an online column-set pool from generator rollouts and derives \textbf{adaptive on-policy retrieval targets} from the column set most frequently associated with execution-correct rollouts. This induces \textbf{bidirectional adaptation}, where the retriever adapts toward column sets that the generator can execute correctly, while the generator adapts to the retriever's evolving schema selections under execution feedback.
With approximately 3k synthetic Text-to-SQL question-database pairs for RL training, ACE-SQL achieves 65.3\% greedy execution accuracy on BIRD Dev while using \textbf{0.93k} output tokens per query.
The repository is available at \url{https://github.com/xbchen1/ACE-SQL}.
\end{abstract}

\section{Introduction}
\begin{figure}[t]
\centering
\includegraphics[width=\columnwidth]{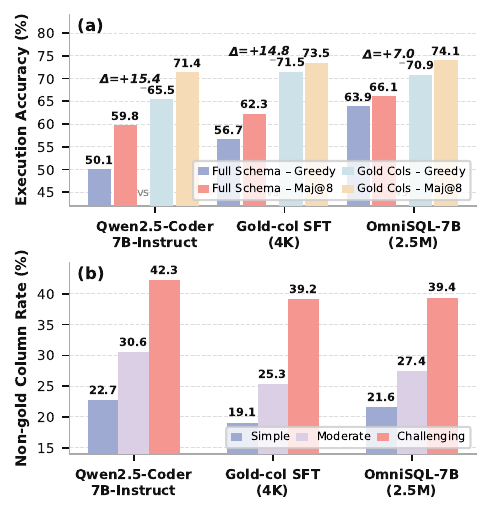}
\caption{Schema retrieval bottleneck on BIRD Dev.
(a) Greedy Execution Accuracy with full-schema vs.\ gold-column inputs.
(b) Rate of non-gold column usage in correct predictions, by difficulty.
Gold-col.\ SFT denotes a Qwen2.5-Coder-7B model lightly fine-tuned on $\sim$4k samples using gold-column inputs.}
\label{fig:pilot}
\end{figure}

Text-to-SQL converts natural language questions into executable SQL queries \citep{deng2022recentadvancestexttosqlsurvey, hong2025nextgenerationdatabaseinterfacessurvey}. Recent LLM-based systems have improved SQL generation through prompting \citep{pourreza2023dinsqldecomposedincontextlearning, dong2023c3zeroshottexttosqlchatgpt, wang2025macsqlmultiagentcollaborativeframework}, supervised fine-tuning \citep{li2024codes, pourreza2024dtssqldecomposedtexttosqlsmall}, and reinforcement learning \citep{ma2026sql}.
As modern databases often contain large and complex schemas, selecting relevant tables and columns becomes a necessary intermediate step for accurate SQL generation, making \textbf{schema linking} a critical challenge~\citep{trust}.
Existing systems typically handle this step in two ways. One approach performs \textbf{full-schema generation}, where schema selection is implicitly learned within end-to-end SQL generation over the entire database \citep{dong2023c3zeroshottexttosqlchatgpt, wang2025macsqlmultiagentcollaborativeframework, lee2024mcssqlleveragingmultipleprompts}. Another approach introduces a separate retriever trained with \textbf{static gold-column supervision} to explicitly select relevant schema components before SQL generation \citep{pourreza2024dtssqldecomposedtexttosqlsmall, glass2025extractive, song2025jolt}.

A central issue in both designs is a \textbf{generator-conditioned supervision mismatch}. In full-schema generation, schema selection is implicitly coupled with SQL generation and updated only through the final SQL loss. As Figure~\ref{fig:pilot}(a) shows, replacing full schemas with gold-column inputs improves execution accuracy by up to \textbf{+15.4} points. Even a model trained on only 4k gold-column samples can \textbf{surpass} a 2.5M-sample model under gold-column inputs, confirming the value of explicit schema linking. In retriever-generator pipelines, schema linking is explicit, but the retriever is trained against static gold-column annotations. Yet Figure~\ref{fig:pilot}(b) shows that from \textbf{19.1\% to 42.3\%} of execution-correct full-schema predictions rely on non-gold column sets, particularly on harder queries. Since SQL generation often admits \textbf{multiple executable routes}, the current generator policy may prefer relational paths and query patterns it has already learned to use. Appendix~\ref{sec:appendix_case_study} provides examples of execution-correct non-gold routes. A retriever anchored to fixed gold targets thus penalizes executable schema configurations preferred by the current generator. These observations suggest that schema linking should be \textbf{explicitly optimized}, and that \textbf{execution-aligned signals} can better track on-policy executable routes.

At the same time, moving retrieval supervision on-policy introduces a \textbf{coupled optimization challenge}: retrieved schemas define the generator's input distribution, while execution-correct generator rollouts determine which column sets become positive retrieval targets. Updating either role can therefore shift the other's training environment or supervision target, creating a circular dependency whose shared backbone further risks gradient interference. However, this same coupling enables \textbf{bidirectional adaptation}: the retriever updates toward column sets that the generator can execute correctly, while the generator adapts to the retriever's evolving selections, with execution accuracy grounding both directions. The challenge is therefore not to eliminate the coupling, but to \textbf{stabilize} it.

We propose \textbf{A}daptive \textbf{C}o-optimization via \textbf{E}mpirical Credit Assignment for Text-to-\textbf{SQL} (\textbf{ACE-SQL}), a reinforcement learning framework that \textbf{jointly optimizes} schema retrieval and SQL generation over a shared policy. ACE-SQL maintains a per-question pool of execution-correct column sets, uses the most frequent set as an \textbf{adaptive on-policy retrieval target}, and trains the generator with execution rewards under a majority-voted schema. PCGrad \citep{yu2020gradient} and a generator-weight schedule stabilize this coupled optimization.

Our contributions are:
\begin{itemize}[leftmargin=*,itemsep=2pt,topsep=2pt]
    \item We formulate schema retrieval in retriever-generator Text-to-SQL as an \textbf{on-policy credit assignment} problem. Instead of relying on static gold-column supervision, ACE-SQL derives adaptive retrieval targets from execution-correct generator rollouts under the current policy, leading to stronger and more stable training.
    \item We propose \textbf{ACE-SQL}, a \textbf{bidirectionally adaptive joint reinforcement learning framework} that uses execution accuracy to align the two directions of the co-adaptation loop. PCGrad and a generator-weight schedule stabilize this coupled optimization.
    \item With only \textbf{2,913} RL examples, ACE-SQL achieves \textbf{65.3\% greedy} execution accuracy on BIRD Dev while using \textbf{0.93k} output tokens per query, outperforming SQL-R1-7B and MTIR-SQL-8B while using 3.3$\times$ and 2.2$\times$ fewer output tokens, and remains competitive on Spider.
\end{itemize}
\section{Related Work}

\paragraph{LLM-Based Text-to-SQL.}
Recent Text-to-SQL systems rely on large language models through prompting \citep{pourreza2023dinsqldecomposedincontextlearning, gao2023texttosqlempoweredlargelanguage, wang2025macsqlmultiagentcollaborativeframework, lee2024mcssqlleveragingmultipleprompts}, supervised fine-tuning \citep{li2024codes, pourreza2024dtssqldecomposedtexttosqlsmall, yang2024synthesizing, li2025omnisql}, and reinforcement learning \citep{ma2026sql}. These methods substantially advance SQL generation, but treat schema linking as either implicit within full-schema generation or frozen within a fixed retriever-generator pipeline. Neither design allows retrieval decisions to receive execution-grounded credit during generator optimization.

\paragraph{Schema Linking in Text-to-SQL.}
Schema linking selects relevant tables and columns to reduce context noise, especially for large databases. Existing approaches include prompt-based schema refinement \citep{cao2024rslsqlrobustschemalinking, lee2024mcssqlleveragingmultipleprompts}, generative pruning \citep{pourreza2024dtssqldecomposedtexttosqlsmall}, extractive or discriminative linking \citep{glass2025extractive, song2025jolt}, and pipeline-based linking with SQL revision \citep{sheng2025base}. Among these, JOLT-SQL \citep{song2025jolt} is the most closely related, as it jointly trains schema linking and SQL generation under a unified objective. However, its retriever target remains the static gold column set throughout training. When the generator's executable preferences diverge from the gold route, as observed in 19\% to 42\% of correct predictions (Figure~\ref{fig:pilot}(b)), retriever updates anchored to a fixed label cannot track this divergence. ACE-SQL instead derives retriever targets from online execution-correct rollouts, allowing the retrieval objective to co-evolve with the generator policy.

\paragraph{Reinforcement Learning for Text-to-SQL.}
Reinforcement learning has been applied to Text-to-SQL through execution rewards for structured query generation \citep{zhong2017seq2sqlgeneratingstructuredqueries}, group-relative policy optimization \citep{shao2024deepseekmath, guo2025deepseek}, task-specific SQL reasoning rewards \citep{ma2026sql, yao2025arctictext2sql}, and multi-turn execution feedback \citep{hua2026sqltrail, xu2025mtirsql}. These systems optimize SQL generation under a fixed or implicit schema context, leaving the retriever outside the RL loop. ACE-SQL instead performs joint RL over both roles: retrieved schemas define the generator's input space, while execution-correct generator rollouts define positive retrieval targets. This bidirectional coupling introduces non-stationarity and gradient conflicts reminiscent of challenges studied in multi-agent RL \citep{foerster2017stabilising} and multi-task optimization \citep{yu2020gradient}. ACE-SQL addresses both with empirical target smoothing, majority voting, a generator-weight schedule, and PCGrad.

\section{ACE-SQL}

\begin{figure*}[t]
\centering
\includegraphics[width=\textwidth]{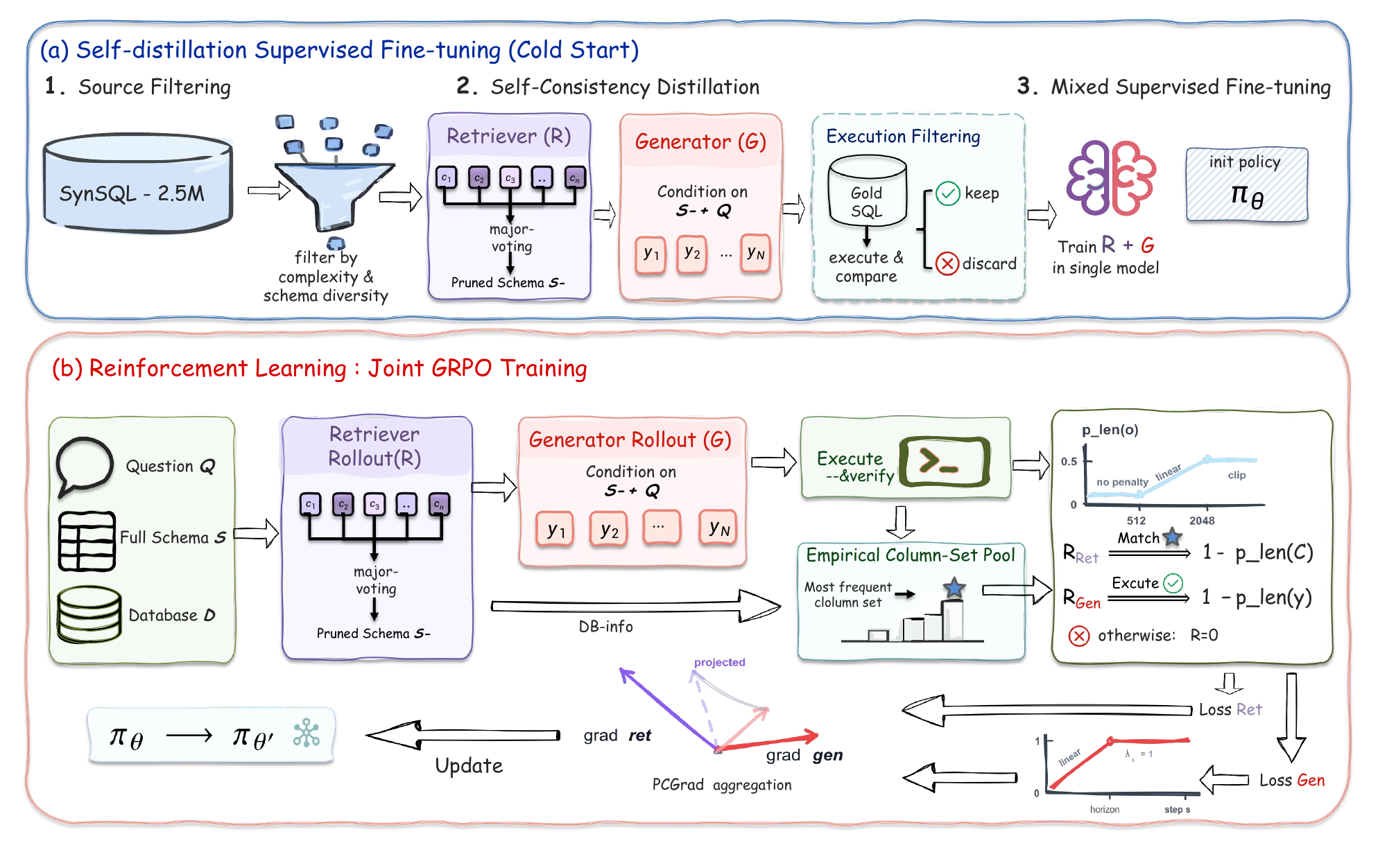}
\caption{Overview of ACE-SQL. \textbf{(a)} Supervised fine-tuning constructs a cold-start retriever-generator pipeline with self-consistency voting and execution-based filtering. \textbf{(b)} Reinforcement learning jointly optimizes the two roles. Correct generator rollouts update an empirical pool of execution-correct column sets, whose most frequent entry is used as the retriever target. Both the retriever and the generator receive rewards and the clipped length penalty only when their outputs match the corresponding targets. The shared policy update uses PCGrad and a generator-weight schedule to stabilize training.}
\label{fig:framework}
\end{figure*}

\subsection{Overview}

Given a natural language question $q$, a database schema $\mathcal{S} = \{(t_i, \{c_{i,j}\})\}$ containing tables $t_i$ with columns $c_{i,j}$, and a database instance $\mathcal{D}$, Text-to-SQL aims to generate an executable SQL query $y$ whose execution result matches the user's intent. ACE-SQL makes \textbf{schema} \textbf{linking} an \textbf{explicit} upstream decision by factorizing inference into two roles handled by the same LLM policy $\pi_\theta$:
\begin{itemize}[leftmargin=*,itemsep=1pt,topsep=2pt]
    \item \textbf{Schema} \textbf{Retrieval}: Given the complete schema $\mathcal{S}$ and question $q$, select a relevant column subset $\hat{\mathcal{C}} \subseteq \mathcal{C}$, where $\mathcal{C} = \{c_{i,j}\}$ denotes all columns.
    \item \textbf{SQL} \textbf{Generation}: Given the pruned schema $\mathcal{S}|_{\hat{\mathcal{C}}}$ and question $q$, generate SQL query $y$.
\end{itemize}

This factorization separates the decisions without separating the backbone. The retriever and generator share parameters, but use different prompts, rollouts, and rewards. As shown in Figure~\ref{fig:framework}, ACE-SQL is trained in two stages. Supervised fine-tuning first provides a cold start for the explicit retriever$\to$generator pipeline (\S\ref{sec:sft}). Joint GRPO training then uses online execution-correct generator rollouts to construct adaptive retriever targets and optimize both roles under the same execution objective (\S\ref{sec:training}). The resulting training dynamics induce \textbf{bidirectional} \textbf{adaptation}: the empirical pool mechanism (\S\ref{sec:training_step}) adapts retriever targets toward generator-preferred executable routes, while majority-voted schema selection adapts the generator's input distribution toward the retriever's evolving consensus. Execution accuracy grounds both directions of this co-adaptation loop.

\subsection{Supervised Fine-Tuning Cold Start}
\label{sec:sft}

Reinforcement learning requires a workable retriever-generator pipeline before \textbf{execution} \textbf{feedback} can be useful. We therefore initialize Qwen3-8B \citep{yang2025qwen3} with \textbf{self-distillation} supervised fine-tuning. This stage is not used to define a permanent gold retrieval target; it only gives both roles enough initial competence for \textbf{online} rollouts to produce execution-matched SQL.

\textbf{(1)} \textbf{Source} \textbf{Filtering}. We first filter samples of suitable difficulty from SynSQL-2.5M \citep{yang2024synthesizing} according to \textbf{query} \textbf{complexity} and \textbf{schema} \textbf{diversity}.

\textbf{(2)} \textbf{Paired} \textbf{Self-Distillation}. We use Qwen3-8B itself to synthesize both retriever and generator samples. For each question, retriever outputs are aggregated with \textbf{self-consistency} voting to obtain candidate column sets. These column sets define \textbf{pruned} schemas for generator sampling, and generator samples are kept only when their \textbf{execution} results match those of the gold SQL\@. If a retriever output cannot support any matched downstream SQL, the corresponding paired sample is discarded.

\textbf{(3)} \textbf{Mixed} \textbf{Supervised} \textbf{Fine-Tuning}. We train the retriever and generator samples together in one full-parameter supervised fine-tuning run. The resulting checkpoint initializes the \textbf{shared} policy $\pi_\theta$ for \textbf{joint} reinforcement learning.

\subsection{Joint GRPO Training}
\label{sec:training}

\subsubsection{Rollout and Empirical Pool}
\label{sec:training_step}

The core of ACE-SQL is to turn successful \textbf{generator} \textbf{behavior} into \textbf{retriever} \textbf{supervision}. Each training-step rollout contains three operations: retriever sampling and parsing, generator sampling and execution, and empirical target update.

\paragraph{Retriever Rollout and Schema Voting.} For each question $q_i$, the policy in retriever mode samples $N$ column selections $\{\hat{\mathcal{C}}_i^{(k)}\}_{k=1}^{N}$. These samples are aggregated by majority voting:
\begin{equation}
    \hat{\mathcal{C}}_i^{\text{maj}} =
    \operatorname{MajVote}\!\left(\{\hat{\mathcal{C}}_i^{(k)}\}_{k=1}^{N}\right).
\end{equation}
A \textbf{pruned} schema containing only columns in $\hat{\mathcal{C}}_i^{\text{maj}}$ is then shared by all generator samples for this question. We denote it as $\hat{\mathcal{S}}_i=\mathcal{S}_i|_{\hat{\mathcal{C}}_i^{\text{maj}}}$. \textbf{Majority} voting reduces \textbf{rollout} noise and provides a stable, \textbf{retriever-conditioned} schema environment for the generator, constituting one direction of \textbf{bidirectional} adaptation: the generator progressively adapts to the schema distribution preferred by the current retriever policy.

\paragraph{Generator Rollout.} The same policy in generator mode produces $N$ responses $\{o_i^{\text{gen},(j)}\}_{j=1}^{N}$ conditioned on the \textbf{pruned} schema, from which SQLs $\{y_i^{(j)}\}_{j=1}^{N}$ are parsed. These SQLs are executed against the database. \textbf{Execution} results provide the generator reward and determine which column sets can be credited to the retriever.

\paragraph{Empirical Column-Set Pool.} The \textbf{empirical} pool records column sets that the generator has successfully used. Before reinforcement learning, we initialize the pool with a \textbf{high-recall} schema rollout from the SFT checkpoint on SynSQL examples labeled as hard. For each question, the retriever samples 8 column selections, and we take their set \textbf{union} rather than selecting a single voted set:
\begin{equation}
    \hat{\mathcal{C}}_{i}^{\text{union}} =
    \bigcup_{k=1}^{8}\hat{\mathcal{C}}_{i,\text{init}}^{(k)}, \qquad
    \hat{\mathcal{S}}_{i}^{\text{init}}=\mathcal{S}_{i}|_{\hat{\mathcal{C}}_{i}^{\text{union}}}.
    \label{eq:init_union_schema}
\end{equation}
The downstream generator then samples 16 SQLs under $\hat{\mathcal{S}}_{i}^{\text{init}}$. This union-based initialization favors \textbf{recall} over early retriever consensus, giving downstream SQL \textbf{exploration} enough schema coverage before the pool is converted into retrieval targets. We retain examples with between 2 and 14 \textbf{execution-correct} SQLs and initialize $f_{\text{pool}}^q$ from the matched initialization SQLs. During training, later matched rollouts continue updating the pool, while old column sets are \textbf{exponentially} decayed so that newer evidence receives sufficient weight. For each \textbf{online} rollout, we extract column sets from matched SQL samples:
\begin{equation}
    \mathcal{P}_{\text{cur}}(q)=
    \{C(y^{(j)}) \mid \text{exec}(y^{(j)})=\text{Match}\},
\end{equation}
where $C(\cdot)$ extracts columns referenced by a SQL query. Let $f_{\text{current}}^q(S)$ count how often a column set $S$ appears in $\mathcal{P}_{\text{cur}}(q)$. These counts update the historical empirical pool for every set $S$ observed in either the current rollout or the existing pool:
\begin{equation}
    f_{\text{pool}}^q(S) \leftarrow \gamma \cdot f_{\text{pool}}^q(S) + f_{\text{current}}^q(S),
\end{equation}
where $\gamma = 0.5$ discounts older rollout evidence while preserving \textbf{successful} routes across updates. Let
\begin{equation}
    S^\star(q)=\arg\max_{S} f_{\text{pool}}^{q}(S)
\end{equation}
be the most frequent successful column set for question $q$. We use $S^\star(q)$ as the \textbf{retriever} \textbf{target}.

\subsubsection{Reward Construction}
\label{sec:reward}

Both roles use \textbf{sparse} rewards gated by the same \textbf{clipped} length-penalty function:
\begin{equation}
    p_{\ell}(o)=0.5\cdot
    \operatorname{clip}\!\left(\frac{\operatorname{len}(o)-512}{2048-512},\,0,\,1\right),
    \label{eq:length_penalty}
\end{equation}
where $\operatorname{len}(\cdot)$ returns the token length and
$\operatorname{clip}(x,0,1)=\min(1,\max(0,x))$. Outputs with at most 512 tokens receive no length penalty, the penalty increases linearly until it reaches 0.5 at 2048 tokens, and outputs are \textbf{truncated} at this maximum response length.

For each question $q$, the generator and retriever rewards are
\begin{align}
r_{\text{gen}}(y)
&=
    \begin{cases}
    1-p_{\ell}(o), & \text{exec}(y)=\text{Match},\\
    0, & \text{otherwise,}
    \end{cases}
\label{eq:gen_reward}\\
r_{\text{ret}}(\hat{\mathcal{C}})
&=
    \begin{cases}
    1-p_{\ell}(o), & \hat{\mathcal{C}}=S^\star(q),\\
    0, & \text{otherwise.}
    \end{cases}
\label{eq:ret_reward}
\end{align}
The SQL $y$ and column set $\hat{\mathcal{C}}$ are parsed from the corresponding role outputs. Thus, the length penalty affects only outputs that satisfy the sparse reward condition. \textbf{Unmatched} outputs receive \textbf{zero} reward regardless of \textbf{length}, preventing reward hacking through short but incorrect responses.

\subsubsection{Joint GRPO Update}
\label{sec:formulation}

\paragraph{Standard GRPO.} GRPO \citep{shao2024deepseekmath} eliminates the value function in PPO \citep{schulman2017proximal} by estimating \textbf{advantages} from within-group reward statistics. We use the standard clipped GRPO surrogate with \textbf{KL} regularization against a frozen reference policy.

\paragraph{Dual-Role Extension.} ACE-SQL applies GRPO to two roles of the same policy. For each question, the retriever and generator each sample $N$ outputs under \textbf{role-specific} prompts. Let $o^{\text{ret},(k)}$ denote the serialized retriever output parsed into $\hat{\mathcal{C}}^{(k)}$, let $o^{\text{gen},(j)}$ denote the serialized generator output parsed into $y^{(j)}$, and write $\hat{\mathcal{S}}$ for the majority-voted pruned schema. The role-specific importance ratios are:
\begin{equation}
\begin{aligned}
\rho^{\text{ret}}_{k,t} &= \frac{\pi_{\theta}(o^{\text{ret}}_{k,t} \mid \mathcal{S}, q, o^{\text{ret}}_{k,<t})}{\pi_{\theta_{\text{old}}}(o^{\text{ret}}_{k,t} \mid \mathcal{S}, q, o^{\text{ret}}_{k,<t})} \\[3pt]
\rho^{\text{gen}}_{j,t} &= \frac{\pi_{\theta}(o^{\text{gen}}_{j,t} \mid \hat{\mathcal{S}}, q, o^{\text{gen}}_{j,<t})}{\pi_{\theta_{\text{old}}}(o^{\text{gen}}_{j,t} \mid \hat{\mathcal{S}}, q, o^{\text{gen}}_{j,<t})}
\end{aligned}
\label{eq:importance_ratios}
\end{equation}
Advantages are normalized within each role's group independently:
\begin{equation}
    \hat{A}^{\text{ret}}_k = \frac{r^{\text{ret}}_k - \mu_{N}^{\text{ret}}}{\sigma_{N}^{\text{ret}}}, \quad \hat{A}^{\text{gen}}_j = \frac{r^{\text{gen}}_j - \mu_{N}^{\text{gen}}}{\sigma_{N}^{\text{gen}}}
    \label{eq:advantages}
\end{equation}
Let $\Phi(o,A;\theta)$ denote the standard per-output GRPO clipped surrogate for a sampled output and its normalized advantage, using the corresponding role-specific importance ratio from Eq.~\ref{eq:importance_ratios}. The role-specific objectives are:
\begin{equation}
\begin{aligned}
\mathcal{J}_{\text{ret}}
&= \mathbb{E}_{q}\frac{1}{N}\sum_{k=1}^{N}
\Phi(o^{\text{ret},(k)}, \hat{A}^{\text{ret}}_k; \theta),\\
\mathcal{J}_{\text{gen}}
&= \mathbb{E}_{q}\frac{1}{N}\sum_{j=1}^{N}
\Phi(o^{\text{gen},(j)}, \hat{A}^{\text{gen}}_j; \theta).
\end{aligned}
\label{eq:dual_role_objectives}
\end{equation}
We use losses $\mathcal{L}_{\text{ret}}=-\mathcal{J}_{\text{ret}}$ and $\mathcal{L}_{\text{gen}}=-\mathcal{J}_{\text{gen}}$. At training step $s$, the generator coefficient $\lambda_s\in[0,1]$ is linearly increased from 0 to 1:
\begin{equation}
\lambda_s = \min\!\left(1,\frac{s-1}{S_\lambda}\right),
\label{eq:lambda_schedule}
\end{equation}
where $S_\lambda$ is the schedule horizon, set to the first 25\% of reinforcement-learning steps in our experiments. \textbf{PCGrad} is applied to role gradients to reduce \textbf{destructive} interference between retriever and generator updates:
\begin{equation}
g_{\text{ACE}}
= \operatorname{PCGrad}\!\left(
g_{\text{ret}},\;
\lambda_s g_{\text{gen}}
\right),
\label{eq:pcgrad_update}
\end{equation}
where $g_{\text{ret}}=\nabla_\theta\mathcal{L}_{\text{ret}}$ and $g_{\text{gen}}=\nabla_\theta\mathcal{L}_{\text{gen}}$. The retriever contributes with weight 1 throughout training, while the generator contribution is \textbf{gradually} activated. This schedule gives the empirical retriever target time to form before generator gradients receive full weight. In implementation, ACE-SQL computes both role losses at each update and applies one optimizer step with the \textbf{projected} joint gradient $g_{\text{ACE}}$.

Algorithm~\ref{alg:ace_sql} in Appendix~\ref{sec:appendix_algorithm} summarizes the reinforcement learning stage of ACE-SQL, including \textbf{online} rollout, \textbf{execution} verification, \textbf{empirical} target update, \textbf{sparse} reward construction, and the PCGrad-based joint update.

\section{Experiments}

\subsection{Setup}

\paragraph{Benchmarks.} We evaluate on \textbf{BIRD} Dev \citep{li2023llmservedatabaseinterface}, which contains 1,534 examples over realistic databases, and on \textbf{Spider} \citep{yu2018spider}. Additional Spider robustness variants, including Spider-DK, Spider-Syn, and Spider-Realistic, are reported in Appendix~\ref{sec:appendix_robustness}.

\paragraph{Training Data.} ACE-SQL uses \textbf{14,184} supervised fine-tuning samples and \textbf{2,913} reinforcement learning question-database pairs. Data construction details are provided in Appendix~\ref{sec:appendix_data}.

\paragraph{Metric.} We report \textbf{greedy} execution accuracy (EX), the percentage of SQL queries generated under greedy decoding whose execution results match the gold SQL\@. Appendix~\ref{sec:appendix_execution} details execution matching and column extraction.

\paragraph{Baselines.} We compare with closed-source prompting systems, including DIN-SQL \citep{pourreza2023dinsqldecomposedincontextlearning}, DAIL-SQL \citep{gao2023texttosqlempoweredlargelanguage}, MAC-SQL \citep{wang2025macsqlmultiagentcollaborativeframework}, and MCS-SQL \citep{lee2024mcssqlleveragingmultipleprompts}. We also compare with open-source Text-to-SQL systems based on base models, supervised fine-tuning, reinforcement learning, and schema-linking modules, including Qwen2.5-Coder \citep{hui2024qwen2}, Qwen3 \citep{yang2025qwen3}, CodeS \citep{li2024codes}, DTS-SQL \citep{pourreza2024dtssqldecomposedtexttosqlsmall}, OmniSQL \citep{li2025omnisql}, SQL-R1 \citep{ma2026sql}, MTIR-SQL \citep{xu2025mtirsql}, JOLT-SQL \citep{song2025jolt}, ExSL \citep{glass2025extractive}, and BASE-SQL \citep{sheng2025base}. The cost analysis additionally reports token usage for MAC-SQL, SQL-R1, and MTIR-SQL.

\paragraph{Implementation.} We train ACE-SQL with \textbf{direct} \textbf{joint} GRPO, initialized from the SFT Qwen3-8B checkpoint \citep{yang2025qwen3}. We set the \textbf{generator-weight} schedule horizon to $S_\lambda=0.25S_{\text{RL}}$, i.e., the generator coefficient reaches 1 after the first 25\% of reinforcement-learning steps. Appendix~\ref{sec:appendix_training} reports optimization and hardware settings.

\subsection{Main Results}

Table~\ref{tab:main_results} presents the main empirical results. Under the greedy decoding setting, \textbf{ACE-SQL} achieves \textbf{65.3\%} execution \textbf{accuracy} on the \textbf{BIRD} development set, outperforming all listed open-source baselines. On Spider, ACE-SQL achieves \textbf{87.2\%} execution accuracy on the test set, while achieving a competitive result of \textbf{79.5\%} on the more challenging Spider-Realistic variant (see Appendix~\ref{sec:appendix_robustness}). For simpler SQL benchmarks characterized by smaller database scales and cleaner schemas, an explicit retrieval stage may occasionally introduce pruning errors without yielding commensurate gains. Conversely, BIRD is closer to real-world database scenarios with larger, noisier, and more complex schemas and queries, where tight \textbf{alignment} between the retriever and the generator becomes far more critical.

\begin{table*}[t]
\centering
\small
\renewcommand{\arraystretch}{0.95}
\setlength{\tabcolsep}{4pt}
\begin{tabular}{@{}
  >{\raggedright\arraybackslash}p{0.20\textwidth}
  >{\raggedright\arraybackslash}p{0.24\textwidth}
  >{\centering\arraybackslash}p{0.13\textwidth}
  >{\centering\arraybackslash}p{0.135\textwidth}
  >{\centering\arraybackslash}p{0.135\textwidth}
  @{}}
\toprule
\rowcolor[HTML]{F5F5F5} \textbf{Method} & \textbf{Base} \textbf{Model} & \textbf{BIRD} \textbf{Dev} & \textbf{Spider} \textbf{Dev} & \textbf{Spider} \textbf{Test} \\
\midrule
\rowcolor[HTML]{FDEEED} \multicolumn{5}{c}{\textit{Closed-Source Large Language Models}} \\
\midrule
DIN-SQL & GPT-4 & 50.7 & 82.8 & 85.3 \\
DAIL-SQL & GPT-4 & 54.8 & 83.6 & 86.6 \\
MAC-SQL & GPT-4 & 59.4 & 86.8 & 82.8 \\
MCS-SQL & GPT-4 & 63.4 & \textbf{89.5} & \textbf{89.6} \\
\midrule
\rowcolor[HTML]{EAF1FC} \multicolumn{5}{c}{\textit{Open-Source Large Language Models}} \\
\midrule
Qwen2.5-Coder & Qwen2.5-Coder-7B & 50.9 & 73.4 & 82.2 \\
DTS-SQL & DeepSeek-Coder-7B & 55.8 & 85.5 & 84.4 \\
CodeS & StarCoderBase-7B & 57.2 & 85.4 & 83.5 \\
JOLT-SQL & Qwen2.5-Coder-7B & 60.4 & 87.0 & 86.8 \\
ExSL & DeepSeek-Coder-7B & 63.2 & 82.4 & 83.0 \\
SQL-R1 & Qwen2.5-Coder-7B & 63.7 & \underline{87.6} & \underline{88.7} \\
BASE-SQL & Qwen2.5-Coder-14B & 63.8 & 86.8 & 87.9 \\
OmniSQL & Qwen2.5-Coder-7B & \underline{63.9} & 81.2 & 87.9 \\
MTIR-SQL & Qwen3-8B & 63.6 & 83.6 & 83.4 \\
\midrule
\textbf{ACE-SQL} & Qwen3-8B & \textbf{65.3} & 83.4 & 87.2 \\
\bottomrule
\end{tabular}
\caption{Main results on BIRD and Spider benchmarks. All result columns report greedy Execution Accuracy (\%). Base Model denotes the backbone used by each method. \textbf{Bold}: highest score; \underline{underline}: second-highest score.}
\label{tab:main_results}
\end{table*}

\subsection{Stabilizers Are Necessary for Joint RL Training}
\label{sec:ablation}

\begin{table}[H]
\centering
\small
\renewcommand{\arraystretch}{1.08}
\setlength{\tabcolsep}{3.5pt}
\begin{tabularx}{0.92\columnwidth}{@{}>{\raggedright\arraybackslash}Xccc@{}}
\toprule
\rowcolor[HTML]{F5F5F5} \textbf{Method} & \textbf{BIRD} & \textbf{Gen.} & \textbf{Ret.} \\
\midrule
Qwen3-8B (base)              & 54.2          & 67.9          & 53.1          \\
\quad + SFT                  & 63.6          & 70.9          & 61.5          \\
\quad + \textbf{ACE-SQL} w/o PCGrad   & 63.2          & 69.2          & 60.7          \\
\quad + \textbf{ACE-SQL} w/o schedule & 64.5          & 71.3          & 63.9          \\
\midrule
\rowcolor[HTML]{EAF1FC}
\quad + \textbf{ACE-SQL} (full)       & \textbf{65.3} & \textbf{72.9} & \textbf{64.2} \\
\bottomrule
\end{tabularx}
\caption{Stabilizer ablation on BIRD Dev (EX, \%).
Gen.\ denotes greedy EX with gold-column inputs; Ret.\ denotes greedy EX when retrieved columns are used as inputs to OmniSQL-7B.}
\label{tab:ablation}
\end{table}
Table~\ref{tab:ablation} isolates the effects of training stages and stabilizers. \textbf{Supervised} fine-tuning raises BIRD Dev from \textbf{54.2\%} to \textbf{63.6\%} and improves retriever ability from \textbf{53.1\%} to \textbf{61.5\%}, providing the necessary \textbf{cold-start} for execution-based reinforcement learning.

The reinforcement learning variants further show why \textbf{joint} \textbf{optimization} requires both \textbf{stabilizers}. Removing \textbf{PCGrad} drops BIRD Dev to 63.2\% and reduces both generator and retriever ability, which is consistent with \textbf{gradient} interference between the two role losses (Appendix~\ref{sec:appendix_gradient_interference}) and indicates partial training collapse. Removing the generator-weight schedule still improves over the SFT baseline, but remains below full ACE-SQL\@. This suggests that, during early training, generator signals can be noisy or even harmful when the generator is optimized under an unstable retriever-defined schema environment. With both stabilizers, ACE-SQL reaches \textbf{65.3\%} BIRD Dev, \textbf{72.9\%} generator ability, and \textbf{64.2\%} retriever ability.

\section{Analysis}
\label{sec:analysis}

\subsection{Gold Targets Are Valid But May Be Suboptimal}

\begin{figure}[!htbp]
\centering
\includegraphics[width=\columnwidth]{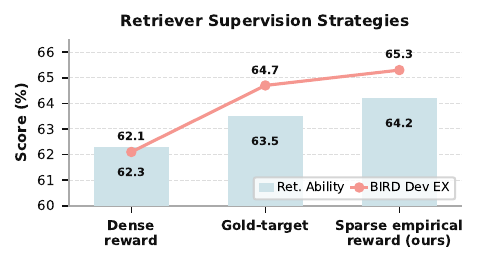}
\caption{Effect of retriever supervision strategies on BIRD Dev. Lines compare pipeline Execution Accuracy and retriever ability (\%).}
\label{fig:reward_strategy}
\end{figure}

\begin{figure}[t]
\centering
\includegraphics[width=\columnwidth]{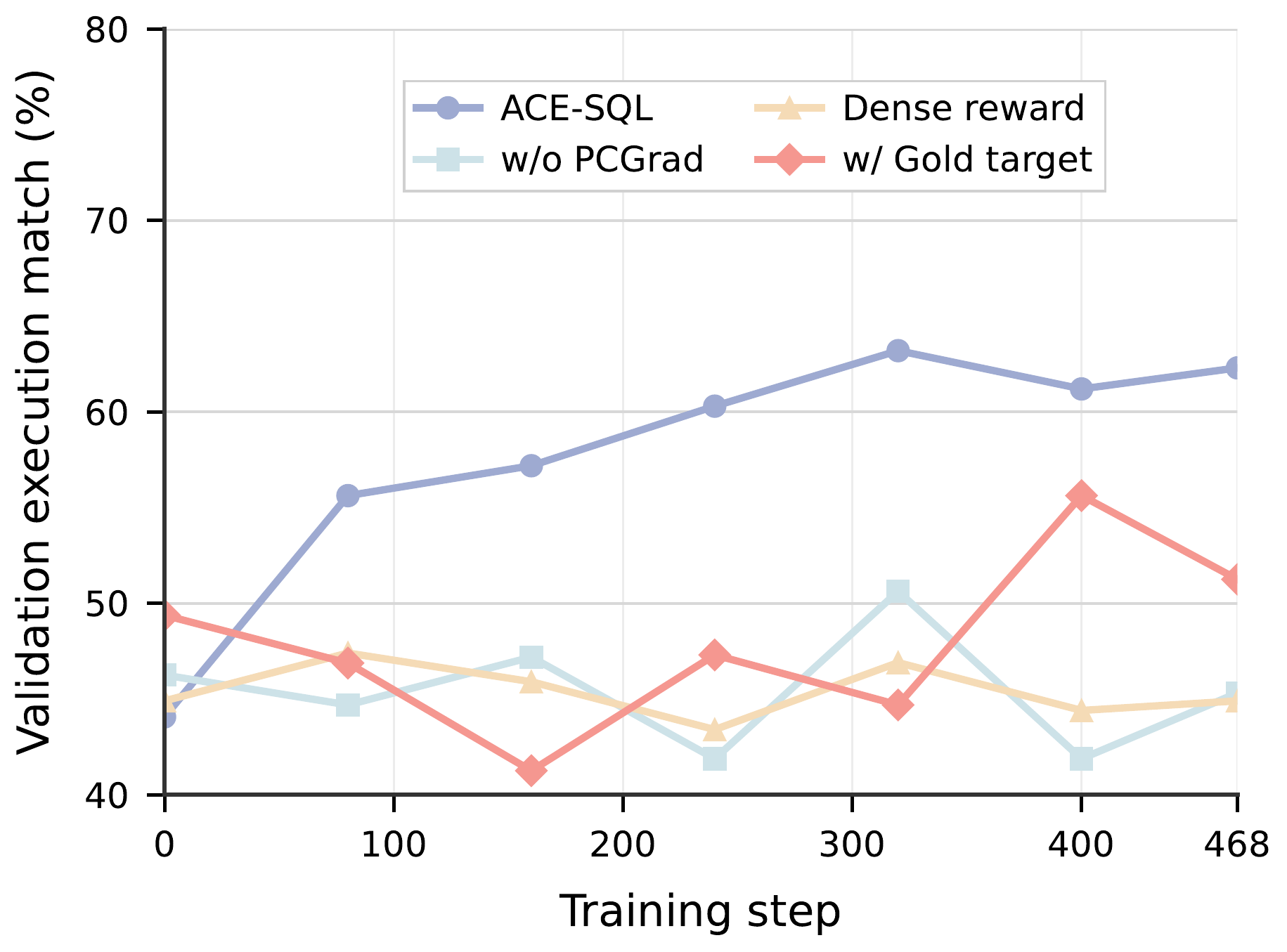}
\caption{Validation EX curves on the held-out validation set over the full training run, evaluated with temperature 0.8 and majority voting over $n=8$ samples.}
\label{fig:reward_training_curves}
\end{figure}

As shown in Figure~\ref{fig:reward_strategy} and Figure~\ref{fig:reward_training_curves}, holding other configurations fixed, using \textbf{gold} columns as the retriever supervision target is a viable training strategy. However, its validation curve shows strong \textbf{instability} and large \textbf{fluctuations} in the early stage, and only begins to generalize clearly on the validation set near the end of training. Its final performance also remains 0.6 points below ACE-SQL.

\begin{table}[!t]
\centering
\small
\renewcommand{\arraystretch}{1.08}
\setlength{\tabcolsep}{3.0pt}
\begin{tabularx}{0.98\columnwidth}{@{}>{\raggedright\arraybackslash}Xcc@{}}
\toprule
\rowcolor[HTML]{F5F5F5}
\textbf{Method} & \textbf{Non-gold} (\%) & \textbf{Hard} \textbf{EX} (\%) \\
\midrule

Qwen3-8B (single-stage)  & 41.2          & 28.1 \\
Qwen3-8B (two-stage)    & 35.7          & 31.5 \\
OmniSQL-7B                  & 39.3          & 47.9 \\
\midrule
\textbf{ACE-SQL} w/ gold target   & 32.5          & 53.1 \\

\rowcolor[HTML]{EAF1FC}
\textbf{ACE-SQL}         & \textbf{45.1} & \textbf{57.2} \\
\bottomrule
\end{tabularx}
\caption{BIRD Dev hard subset analysis at $n{=}8$, temperature $= 0.8$. Non-gold reports rate of non-gold column usage in
correct predictions; Hard EX reports execution accuracy on the hard subset of BIRD Dev.}
\label{tab:hard_non_gold_columns}
\end{table}

This indicates that although the gold target is valid, it is not always the \textbf{best} target for the current \textbf{policy}. As the retriever is optimized toward static gold-column selections, the generator is increasingly exposed to a schema distribution that may differ from the executable paths it has learned to use, making early training more brittle. In contrast, \textbf{execution-based} credit assignment turns this one-way dependency into a closed, mutually adaptive loop.

Table~\ref{tab:hard_non_gold_columns} supports this on the BIRD Dev hard subset ($n{=}8$). ACE-SQL w/ gold target yields the lowest \textbf{non-gold} ratio (32.5\%), confirming that static supervision constrains the retriever to annotated routes, yet its hard EX falls 4.1 points behind full ACE-SQL. Moreover, despite the two-stage pipeline inherently limiting column exploration through explicit schema pruning, ACE-SQL still exhibits the highest non-gold ratio (45.1\%) among all methods while achieving the highest hard EX. This indicates that \textbf{empirical} credit assignment captures the policy's own preferred \textbf{executable} routes rather than imposing a fixed annotated path, allowing the model to leverage its learned SQL reasoning patterns for more robust performance.

\subsection{Sparse Rewards Perform Better}
We compare two empirical rewards for the retriever: a \textbf{sparse} reward and a \textbf{dense} reward. The sparse reward, defined in Section~\ref{sec:reward}, is given only when the selected column set exactly matches the empirical target $S^\star(q)$, while the dense reward assigns a shaped score using continuous soft signals derived from empirical-pool coverage and noise ratio, as detailed in Appendix~\ref{sec:appendix_reward}. As shown in Figure~\ref{fig:reward_strategy} and Figure~\ref{fig:reward_training_curves}, the dense reward improves retriever ability by only \textbf{+0.8} points, much less than the \textbf{+2.7} points achieved by the sparse reward, and even degrades the full pipeline by \textbf{-1.5} points relative to the SFT starting point.

This may be because the dense reward can favor locally reasonable but incomplete column selections, leading to \textbf{reward} \textbf{hacking}. Moreover, in joint optimization, when the generator reward is sparse, the dense retriever reward can more easily produce larger and more frequent gradients, dominate the \textbf{training} direction, and degrade the generation task, as further discussed in Appendix~\ref{sec:appendix_sparse_reward}.

\subsection{Cost Analysis}

The two-stage retriever-generator pipeline produces both retrieval and generation outputs, so we report the sum of their average generated output \textbf{tokens} on BIRD Dev as a model-side \textbf{cost} proxy. We compare against Qwen3-8B (base) under the same two-stage prompting setup, and against external baselines including MAC-SQL, SQL-R1-7B, and MTIR-SQL\@. As shown in Table~\ref{tab:cost_lengths}, SQL-R1-7B reaches 63.7\% BIRD Dev accuracy with 3.10k output tokens, while ACE-SQL reaches 65.3\% with 0.93k output tokens.

This comparison indicates that ACE-SQL's explicit schema-retrieval stage does \textbf{not} rely on \textbf{longer} generations to obtain its \textbf{gain}. Instead, isolating schema retrieval as an explicit stage and restricting the generator context improve both \textbf{inference} efficiency and prediction quality. Relative to SQL-R1-7B, ACE-SQL uses about \textbf{70\%} fewer output tokens while improving BIRD Dev execution accuracy by \textbf{+1.6} points. Within our own pipeline, reinforcement learning reduces the average generated length from \textbf{1.90k} tokens after supervised fine-tuning to \textbf{0.93k} tokens, reflecting the importance of the gated length penalty.

\begin{center}
\footnotesize
\renewcommand{\arraystretch}{1.08}
\setlength{\tabcolsep}{2.8pt}
\begin{tabularx}{\columnwidth}{@{}>{\raggedright\arraybackslash}p{0.44\columnwidth}>{\centering\arraybackslash}p{0.25\columnwidth}>{\centering\arraybackslash}X@{}}
\toprule
\rowcolor[HTML]{F5F5F5} \textbf{Method} & \mbox{\textbf{BIRD} \textbf{Dev} \textbf{EX}} & \mbox{\textbf{Tokens} \textbf{(k)} $\downarrow$} \\
\midrule
MAC-SQL + GPT-4 & 59.4 & 2.17 \\
SQL-R1-7B    & 63.7 &   3.10 \\
MTIR-SQL-4B  & 63.1 &   2.90 \\
MTIR-SQL-8B  & 63.6 &   2.00 \\
\midrule
Qwen3-8B (base)  & 54.2 &  2.10 \\
\textbf{ACE-SQL} (SFT)    & 63.6 &  1.90 \\
\rowcolor[HTML]{EAF1FC}
\textbf{ACE-SQL} (SFT + RL)   & \textbf{65.3} & \textbf{0.93} \\
\bottomrule
\end{tabularx}
\captionof{table}{Inference cost on BIRD Dev (EX, \%). Tokens: average generated output tokens per query.}
\label{tab:cost_lengths}
\end{center}
\section{Conclusion}

We propose ACE-SQL, a reinforcement learning framework that \textbf{jointly} optimizes schema retrieval and SQL generation through dual-role GRPO with \textbf{empirical} credit assignment. Its core idea is to use \textbf{execution-correct} SQL rollouts as an \textbf{on-policy} basis for assigning credit to explicit schema-retrieval actions, rather than forcing retrieval supervision to follow a single, static gold column set. Joint on-policy training creates \textbf{bidirectional} adaptation, allowing the generator to adapt to the retriever's evolving schema selections and the retriever to adapt to the generator's execution-correct outputs. To address the resulting coupling problem, ACE-SQL stabilizes the optimization process with an empirical column-set pool, \textbf{PCGrad}, and a generator-weight schedule. Execution accuracy provides a \textbf{shared} grounding signal that keeps both directions aligned around correct SQL execution. On BIRD Dev, ACE-SQL reaches \textbf{65.3\%} execution accuracy while reducing average output length to \textbf{0.93k} tokens, suggesting a practical direction for efficient and robust Text-to-SQL systems over complex real-world databases. Moreover, this approach provides an on-policy perspective on general upstream-downstream credit assignment.

\section*{Limitations}

Our work has several limitations. First, all experiments use a single 8B-parameter model (Qwen3-8B), and further scaling experiments on larger models and different architectures would better establish generalizability. However, such experiments require substantially more computational resources and are beyond our available compute budget. Second, training data comes exclusively from synthetic SynSQL-2.5M; real-world query-database pairs may reveal different dynamics. Third, the empirical pool is initialized from the supervised checkpoint rather than from the base model. This improves training efficiency and simplifies the pipeline, but may under-explore executable SQL preferences that the base model could express before supervised schema-retrieval adaptation.

\bibliography{custom}

\appendix

\section{ACE-SQL Algorithm}
\label{sec:appendix_algorithm}

\begin{algorithm}[H]
\footnotesize
\caption{ACE-SQL reinforcement learning with empirical credit assignment}
\label{alg:ace_sql}
\begin{algorithmic}[1]
\Require Policy $\pi_\theta$, RL data $\mathcal{D}_{\text{RL}}$, empirical pools $\{f_{\text{pool}}^q\}$, decay $\gamma$, group size $N$, schedule horizon $S_\lambda$
\For{training step $s=1,2,\ldots$}
    \State Sample a batch from $\mathcal{D}_{\text{RL}}$
    \For{each question $q$ in the batch}
        \State Sample $N$ retriever outputs $\{\hat{\mathcal{C}}^{(k)}\}_{k=1}^{N}$ and aggregate $\hat{\mathcal{C}}^{\text{maj}}$ by majority voting
        \State Sample $N$ generator outputs $\{o^{\text{gen},(j)}\}_{j=1}^{N}$ using $\mathcal{S}|_{\hat{\mathcal{C}}^{\text{maj}}}$ and parse SQLs $\{y^{(j)}\}_{j=1}^{N}$
        \State $\mathcal{P}_{\text{cur}}(q)\leftarrow\{C(y^{(j)}):\textsc{ExecMatch}(y^{(j)},y^\star)\}$
        \State Count $f_{\text{current}}^q(S)$ from $\mathcal{P}_{\text{cur}}(q)$ and update $f_{\text{pool}}^q(S)\leftarrow\gamma f_{\text{pool}}^q(S)+f_{\text{current}}^q(S)$
        \State $S^\star(q)\leftarrow\arg\max_S f_{\text{pool}}^q(S)$
        \State Assign generator rewards by execution match and retriever rewards by exact match to $S^\star(q)$
        \State Apply the shared clipped length penalty $p_\ell(\cdot)$ only to matched outputs in both roles
    \EndFor
    \State Compute role-specific GRPO losses $\mathcal{L}_{\text{ret}}$ and $\mathcal{L}_{\text{gen}}$
    \State $\lambda_s\leftarrow \min(1,(s-1)/S_\lambda)$
    \State $g_{\text{ACE}}\leftarrow \operatorname{PCGrad}(\nabla_\theta\mathcal{L}_{\text{ret}},\lambda_s\nabla_\theta\mathcal{L}_{\text{gen}})$
    \State Update $\theta$ using $g_{\text{ACE}}$
\EndFor
\end{algorithmic}
\end{algorithm}

\section{Data Construction}
\label{sec:appendix_data}

ACE-SQL uses a two-stage training recipe: supervised fine-tuning establishes the explicit retriever$\to$generator pipeline, and reinforcement learning then optimizes the same policy with execution-grounded empirical credit.

\subsection{Training Data Overview}
\label{sec:appendix_data_overview}
\begin{center}
\footnotesize
\renewcommand{\arraystretch}{0.75}
\setlength{\tabcolsep}{3pt}
\begin{tabularx}{\columnwidth}{@{\hspace{\tabcolsep}}
  >{\raggedright\arraybackslash}m{0.27\columnwidth}
  >{\raggedright\arraybackslash}p{0.37\columnwidth}
  >{\scriptsize\raggedright\arraybackslash}X
  @{\hspace{\tabcolsep}}}
\toprule
\rowcolor[HTML]{F5F5F5} \textbf{Model / Setting} & \textbf{SFT Data} & {\footnotesize\textbf{RL Data}} \\
\midrule
Gold-col.\ SFT         & $\sim$4k gold-col.\ samples & N/A \\
OmniSQL               & 2.5M SynSQL                & N/A \\
SQL-R1                & 2.5M SynSQL                & 5k SynSQL \\
MTIR-SQL              & N/A                        & 18.1k Spider+BIRD \\
\textbf{ACE-SQL}               & 14,184 filtered SynSQL samples & 2,913 hard pairs \\
\bottomrule
\end{tabularx}
\captionof{table}{Training data overview for compared models. ``Gold-col.\ SFT'' refers to a preliminary gold-column SFT bottleneck study. OmniSQL, SQL-R1, and MTIR-SQL are external baselines. ACE-SQL is our method.}
\label{tab:training_data}
\end{center}
\subsection{Supervised Fine-Tuning Data}
\label{sec:appendix_sft_data}

For supervised fine-tuning, we filter approximately 9,000 question-database pairs from SynSQL-2.5M \citep{yang2024synthesizing} based on the dataset-provided query-complexity labels and the diversity of the corresponding database schemas. We then apply self-consistency voting ($n=16$) with execution-based filtering to produce 14,184 balanced retriever and generator training samples (7,092 per role). Retriever samples provide structured column selections, and generator samples use the corresponding pruned schemas as input. Samples whose downstream SQL execution does not match the gold SQL are discarded.

\subsection{Reinforcement Learning Data}
\label{sec:appendix_rl_data}

Reinforcement learning uses hard samples from SynSQL-2.5M. We start from the subset of the SFT source data directly labeled as hard in SynSQL, containing 5,637 question-database pairs. Before reinforcement learning, we run an extended rollout with the SFT checkpoint: the retriever samples 8 outputs, their selected columns are unioned into one initialization schema, and the downstream generator samples 16 SQLs under that schema. We retain examples for which between 2 and 14 of the 16 generator rollouts are execution-correct, yielding 2,913 question-database pairs for RL training. The same rollout results initialize a per-question empirical pool. During each RL rollout, retriever samples are aggregated through the majority-voting procedure in Section~\ref{sec:training_step}; generator executions then update the empirical column-set pool with the decay factor $\gamma = 0.5$, and the decayed pool defines the sparse retriever target $S^\star(q)$.

\section{Additional Benchmark Results}
\label{sec:appendix_robustness}

Table~\ref{tab:spider_robustness} reports Spider robustness variants that are omitted from the main table for compactness. We include open-source systems in the 7B to 8B scale with reported values.

\begin{center}
\centering
\scriptsize
\resizebox{\columnwidth}{!}{%
\begin{tabular}{@{}lccc@{}}
\toprule
\rowcolor[HTML]{F5F5F5} \textbf{Model} & \textbf{Spider-DK} & \textbf{Spider-Syn} & \textbf{Spider-Realistic} \\
\midrule
OmniSQL-7B  & \textbf{76.1} & 69.7          & 76.2          \\
MTIR-SQL-8B & 72.9          & \textbf{77.2} & \underline{77.4} \\
\rowcolor[HTML]{EAF1FC} \textbf{ACE-SQL} & \underline{74.4} & \underline{73.7} & \textbf{79.5} \\
\bottomrule
\end{tabular}
}
\captionof{table}{Additional results on Spider robustness variants. All columns report greedy Execution Accuracy.}
\label{tab:spider_robustness}
\end{center}

\section{Training Configuration and Prompts}
\label{sec:appendix_training}

\subsection{Hyperparameters}
\label{sec:appendix_hyperparams}

Table~\ref{tab:hyperparams} summarizes the hyperparameters used in each training stage.
\begin{table*}[t]
\centering
\small
\renewcommand{\arraystretch}{0.95}
\setlength{\tabcolsep}{14pt}
\begin{tabularx}{\textwidth}{@{\hspace{\tabcolsep}}>{\raggedright\arraybackslash}Xcc@{\hspace{\tabcolsep}}}
\toprule
\rowcolor[HTML]{F5F5F5} \textbf{Hyperparameter} & \textbf{Supervised Fine-Tuning} & \textbf{Reinforcement Learning} \\
\midrule
Base model                & Qwen3-8B                       & SFT checkpoint \\
Training paradigm         & Full-parameter tuning          & Joint GRPO \\
Hardware                  & 4$\times$NVIDIA A100 80GB PCIe & 4$\times$NVIDIA A100 80GB PCIe \\
Training framework        & LLaMA-Factory                  & VERL 0.5.5 + vLLM \\
Wall-clock time           & $\sim$38 hours                 & $\sim$43 hours \\
\midrule
Gradient surgery          & N/A                            & PCGrad \\
Loss weights              & N/A                            & Retriever: 1; Generator: $\lambda_s\!:\!0\!\rightarrow\!1$ \\
Schedule horizon $S_\lambda$ & N/A                         & 25\% of RL steps \\
Learning rate             & $2\times10^{-5}$               & $1\times10^{-6}$ \\
Learning-rate warmup      & 5\%                            & 5\% \\
Epochs                    & 2                              & 4 \\
\midrule
Max prompt length         & 4096                           & 4096 \\
Max response length       & 2048                           & 2048 \\
Group size $N$            & N/A                            & 8 \\
Temperature               & N/A                            & 1.0 \\
KL divergence coeff.      & N/A                            & 0.001 \\
\bottomrule
\end{tabularx}
\caption{Hyperparameters for supervised fine-tuning and reinforcement learning stages.}
\label{tab:hyperparams}
\end{table*}
\subsection{Prompt Templates}
\label{sec:appendix_prompts}

We use structured prompts for both the retriever and generator roles. The retriever prompt includes the full database schema (all tables and columns with types) and the natural language question, and instructs the model to output relevant columns in a structured format. The generator prompt includes only the pruned schema (tables and columns selected by the retriever) and the question, and instructs the model to produce SQL within tagged code blocks preceded by reasoning in \texttt{<think>} tags.

\begin{figure*}[t!] 
\centering
\includegraphics[width=0.98\textwidth]{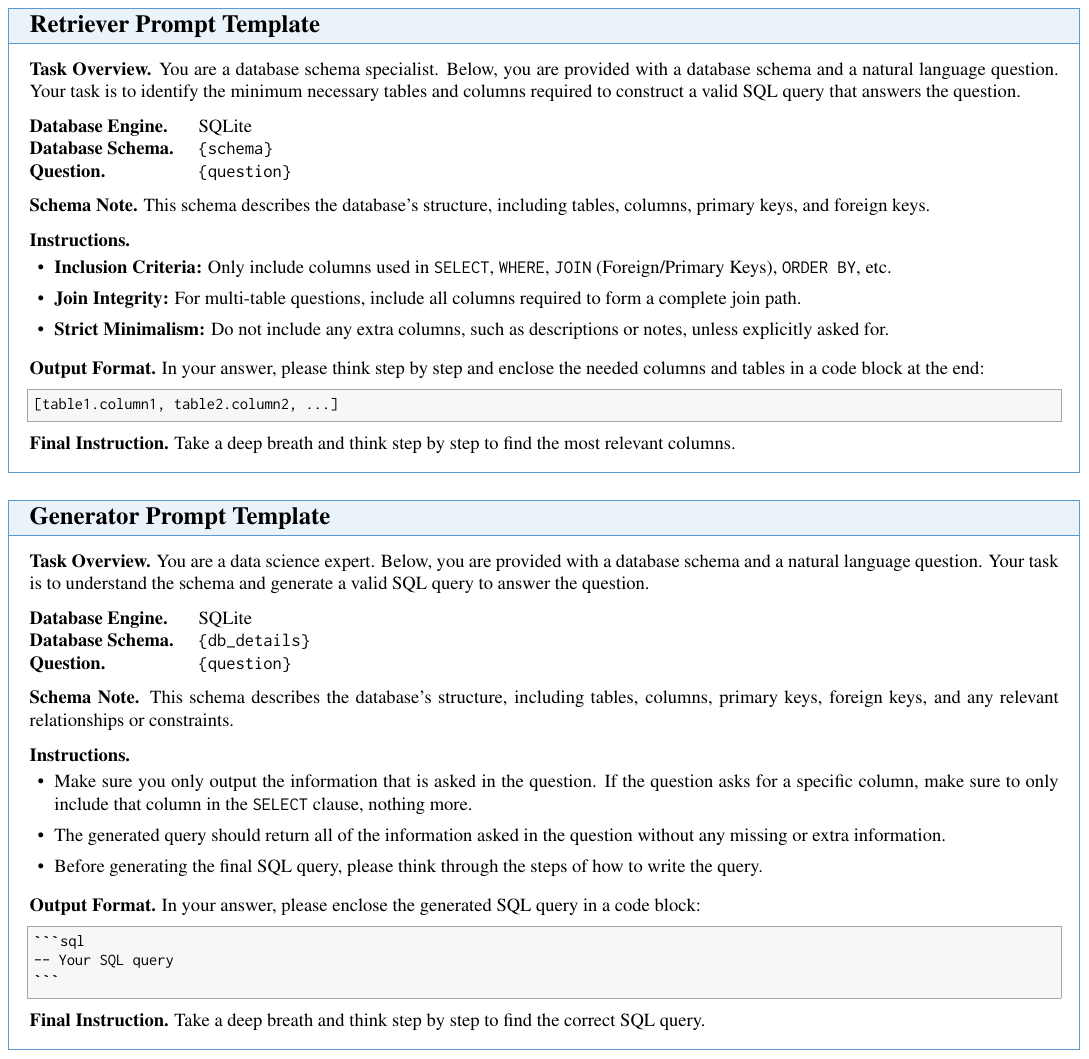}
\caption{Prompt templates for the retriever and generator roles.}
\label{fig:prompt_templates}
\end{figure*}

Figure~\ref{fig:prompt_templates} shows the rendered prompt templates used in our implementation.

\subsection{Execution Matching and Column Extraction}
\label{sec:appendix_execution}

We use execution matching to evaluate generated SQL queries. We reuse the relatively strict execution function from SQL-R1, which is also trained on SynSQL\@. During evaluation, we use greedy decoding with one generated SQL per example and apply the execution code associated with the corresponding benchmark. During training rollouts on SynSQL, we use a 30-second execution timeout because the SynSQL databases are relatively small. During evaluation, we use a 3-minute timeout.

For empirical credit assignment, we extract columns from matched SQL queries with SQLGlot. Extracted table and column names are canonicalized through DB-info, which maps SQL mentions back to database table and column identifiers before updating the empirical column-set pool.

\section{Reward Design Details}
\label{sec:appendix_reward}

\subsection{Dense Retriever Reward Variant}
\label{sec:appendix_sparse_reward}

We compare ACE-SQL with a dense retrieval reward variant to isolate the effect of sparse empirical credit. This variant is a natural shaped reward over the empirical pool: it gives credit when the selected columns cover any execution-correct route already stored in the pool, and penalizes only columns that never appear in the pool. Let $\mathcal{B}^q=\{S:f_{\text{pool}}^q(S)>0\}$ be the support of the empirical pool for question $q$, and let $U^q=\bigcup_{S\in \mathcal{B}^q}S$ be the union of columns appearing in the pool. The dense variant scores a selected column set by pool coverage minus a noise penalty:
\begin{equation}
\begin{aligned}
    r_{\text{dense-ret}}(\mathcal{C})
    &= \operatorname{coverage}(\mathcal{C}, \mathcal{B}^q)\\
    &\quad - 0.5 \cdot \operatorname{noise}(\mathcal{C}, U^q),
\end{aligned}
\end{equation}
where
\begin{equation}
\operatorname{coverage}(\mathcal{C}, \mathcal{B}^q)
    = \max_{S\in \mathcal{B}^q}\frac{|\mathcal{C}\cap S|}{|S|}
\end{equation}
compares the selected set with every column set in the empirical pool and uses the highest coverage ratio. The noise term is
\begin{equation}
\operatorname{noise}(\mathcal{C}, U^q)
    = \frac{|\mathcal{C}\setminus U^q|}{|\mathcal{C}|},
\end{equation}
so only columns completely absent from the pool are counted as noise. The generator reward, PCGrad update, and generator-weight schedule are otherwise unchanged.

Dense retrieval rewards are not uninformative in isolation; they can provide frequent local feedback. The issue is their behavior inside direct joint reinforcement learning with a shared policy. When the retriever reward is dense, the retriever side receives a much more continuous advantage signal than the generator side, whose execution reward is naturally sparse. On difficult samples, this can induce a simple path dependence: the retriever may settle into an intermediate state that looks locally reasonable but is not further refined, while the larger retriever gradients weaken generator learning.

\begin{center}
\begin{minipage}{\columnwidth}
\centering
\includegraphics[width=\linewidth]{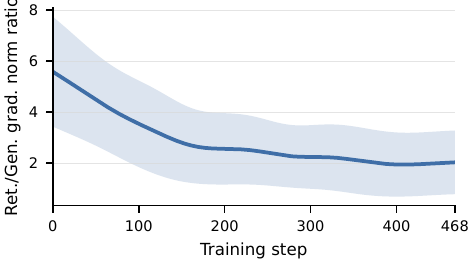}
\captionof{figure}{Retriever-to-generator gradient norm ratio under the dense retriever reward variant.}
\label{fig:dense_reward_gradient_ratio}
\end{minipage}
\end{center}

Figure~\ref{fig:dense_reward_gradient_ratio} further shows that, under the dense retriever reward variant, the retriever gradient magnitude remains larger than the generator gradient magnitude throughout training. In the early stage, this imbalance is partly caused by the generator-weight schedule, but it remains around 2$\times$ for much of the middle and late stages. This result shows that reward-density imbalance can induce a persistent gradient-magnitude imbalance between the two roles.

\begin{table}[tbp]
\centering
\scriptsize
\resizebox{\columnwidth}{!}{%
\begin{tabular}{@{}lcc@{}}
\toprule
\rowcolor[HTML]{F5F5F5} \textbf{Retriever Reward} & \textbf{BIRD Dev} & \textbf{Retriever Ability} \\
\midrule
Dense coverage-noise reward      & 62.1 & 62.3 \\
\rowcolor[HTML]{EAF1FC}
\textbf{ACE-SQL} sparse empirical reward  & \textbf{65.3} & \textbf{64.2} \\
\bottomrule
\end{tabular}
}
\caption{Effect of retriever reward design. ``Retriever Ability'' fixes OmniSQL-7B as the downstream generator and evaluates schema retrieval quality.}
\label{tab:dense_reward}
\end{table}

\section{Additional Analysis Details}
\label{sec:appendix_analysis}

\subsection{Gradient Conflict without PCGrad}
\label{sec:appendix_gradient_interference}

Figure~\ref{fig:pcgrad_conflict_ratio} reports the smoothed conflict ratio between retriever and generator gradients before PCGrad projection. The conflict ratio stays high across training, indicating that the two role losses frequently propose incompatible shared-backbone updates. This persistent conflict explains why directly summing the two role gradients can reduce both generator and retriever ability, and why PCGrad is needed to stabilize joint reinforcement learning.

\begin{center}
\begin{minipage}{\columnwidth}
\centering
\includegraphics[width=\linewidth]{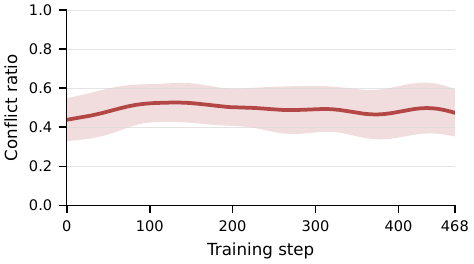}
\captionof{figure}{Gradient conflict ratio between retriever and generator role-loss gradients without PCGrad projection, defined as the proportion of gradient pairs with cosine similarity below zero.}
\label{fig:pcgrad_conflict_ratio}
\end{minipage}
\end{center}

\subsection{Impact of Empirical Targets}
\label{sec:appendix_empirical_gold}

The ACE-SQL gold-target baseline in Figure~\ref{fig:reward_strategy} uses the same training data as ACE-SQL and serves as a complementary reference. It optimizes a correct column set, but the retriever still receives sparse updates anchored to a single annotated route. When that route is far from the current policy's executable path distribution, the optimization trajectory can become less stable and can incur some performance loss even though the target itself is valid. ACE-SQL instead refreshes empirical targets through online rollouts and execution verification, allowing successful non-gold routes to enter the decayed pool that defines retriever supervision.

\subsection{Cost Analysis}
\label{sec:appendix_cost}

For Table~\ref{tab:cost_lengths}, we use output tokens as the shared cost proxy because the latency and tool-call counts of external methods depend on implementation, hardware, and interaction protocol. Our values are measured as the average total generated output length of each model variant on BIRD Dev. For two-stage variants, this total includes both retriever and generator outputs.

\subsection{Case Study}
\label{sec:appendix_case_study}

\begin{figure*}[t!]
\centering
\includegraphics[width=0.98\textwidth]{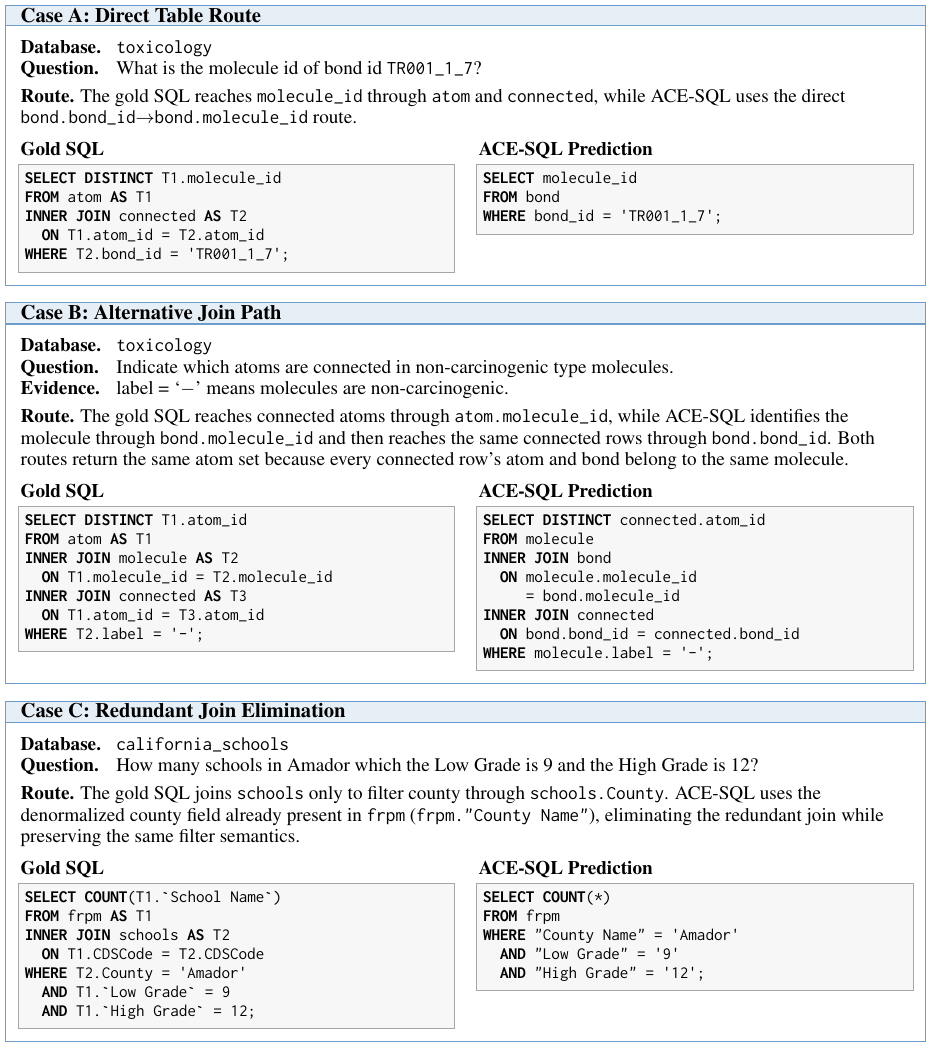}
\caption{Qualitative examples of execution-correct non-gold routes. Each box reports the question, route difference, gold SQL, and ACE-SQL prediction.}
\label{fig:case_study}
\end{figure*}

Figure~\ref{fig:case_study} provides three qualitative examples from BIRD Dev to illustrate why non-gold executable routes can be useful retriever targets. In each case, ACE-SQL uses a column route different from the gold SQL but returns the same execution result under the row- and column-order preserving execution comparison used in our evaluation.

\clearpage

\end{document}